\pdfoutput=1

\documentclass[11pt]{article}

\usepackage[]{ACL2023}

\usepackage{times}
\usepackage{latexsym}
\usepackage{ulem}
\usepackage[T1]{fontenc}

\usepackage[utf8]{inputenc}

\usepackage{microtype}

\usepackage{inconsolata}
\usepackage{booktabs}

\title{
AoM: Detecting Aspect-oriented Information for Multimodal \\
Aspect-Based Sentiment Analysis}

\author{
Ru Zhou\textsuperscript{1}\;\;\;
Wenya Guo\textsuperscript{1}\thanks{\; Corresponding author.}\;\;\; 
Xumeng Liu\textsuperscript{1}\;\;\;
Shenglong Yu\textsuperscript{1}\\
{\bf 
Ying Zhang\textsuperscript{1}\;\;\;
Xiaojie Yuan\textsuperscript{1}}\\
\textsuperscript{1} College of Computer Science, TKLNDST, Nankai University, Tianjin, China \; \\
{\tt \{zhouru,guowenya,liuxumeng,yushenglong,zhangying\}@dbis.nankai.edu.cn}\\
{\tt yuanxj@nankai.edu.cn}}

\usepackage{graphicx}
\usepackage{stfloats}
\usepackage{amsfonts}
\usepackage{amsmath}
\usepackage{multirow} 
\usepackage{bbding}

\begin{document}
\maketitle

\begin{abstract}
%
Multimodal aspect-based sentiment analysis (MABSA) aims to extract aspects from text-image pairs and recognize their sentiments.
%
Existing methods make great efforts to align the whole image to corresponding aspects. 
%
%
However, different regions of the image may relate to different aspects in the same sentence, and coarsely establishing image-aspect alignment will introduce noise to aspect-based sentiment analysis (\textit{i.e.}, visual noise). 
Besides, the sentiment of a specific aspect can also be interfered by descriptions of other aspects (\textit{i.e.}, textual noise).
%
%
%
Considering the aforementioned noises, this paper proposes an \underline{A}spect-\underline{o}riented \underline{M}ethod (\textbf{AoM}) to detect aspect-relevant semantic and sentiment information.
Specifically, an aspect-aware attention module is designed to simultaneously select textual tokens and image blocks that are semantically related to the aspects. 
%
%
To accurately aggregate sentiment information, we explicitly introduce sentiment embedding into AoM, and use a graph convolutional network to model the vision-text and text-text interaction. 
Extensive experiments demonstrate the superiority of AoM to existing methods.
The source code is publicly released at \href{https://github.com/SilyRab/AoM}{https://github.com/SilyRab/AoM}.

%
%
%
%
%
%
\end{abstract}

\section{Introduction}
As an important and promising task in the field of sentiment analysis, 
Multimodal Aspect-Based Sentiment Analysis (MABSA) has attracted increasing attention \citep{lv2021aspect,juJointMultimodalAspectSentiment2021}. 
Given an image and corresponding text, 
MABSA is defined as jointly extracting all aspect terms from image-text pairs and predicting their sentiment polarities \citep{juJointMultimodalAspectSentiment2021}. 
%

\begin{figure}
\centering 
\includegraphics[width = 0.95 \linewidth]{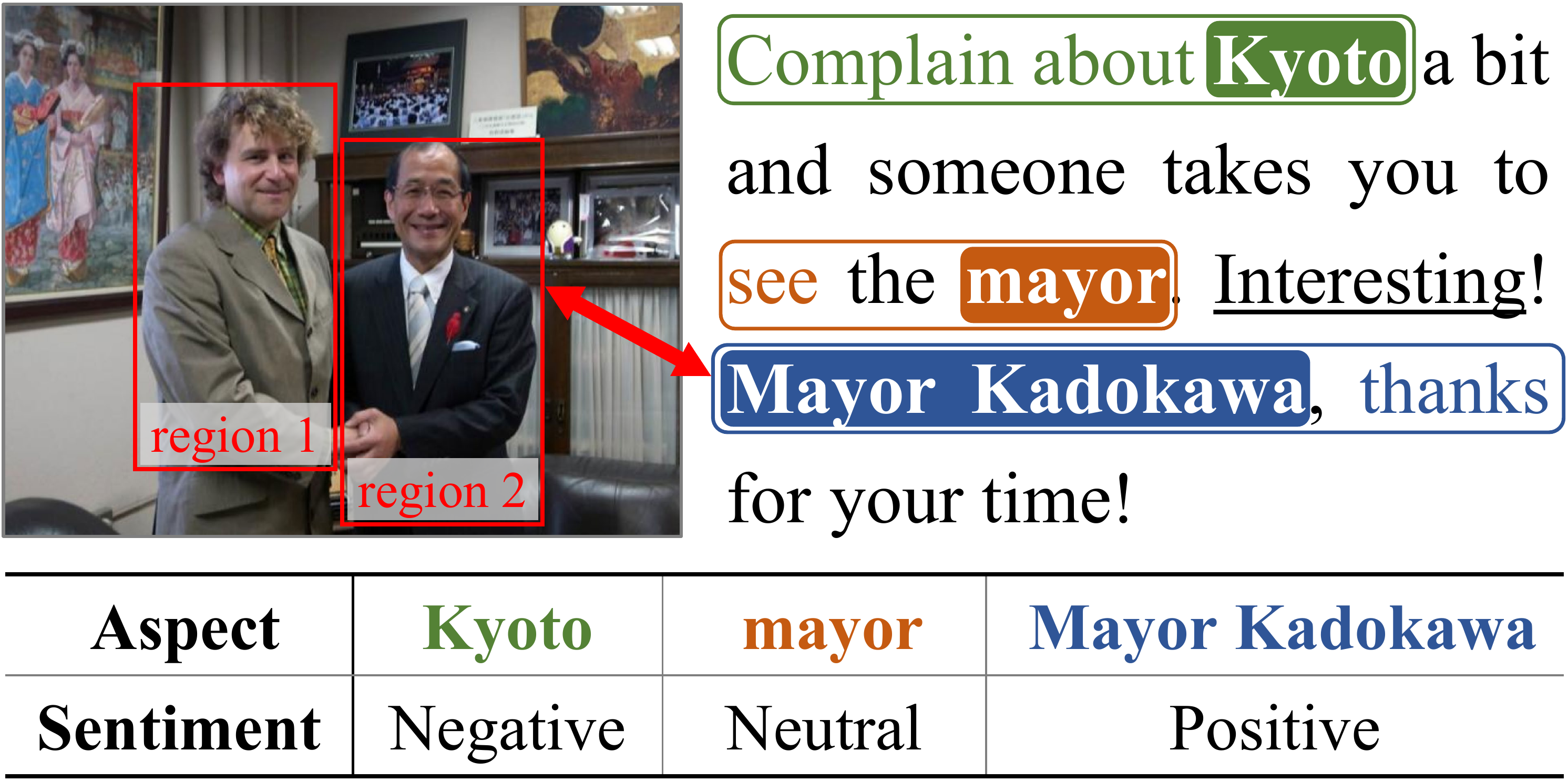}
\caption{
An example of MABSA task, including the aspects, their corresponding descriptions, and sentiments. 
%
}
\label{fig:intro}
\end{figure}

In this scenario of fine-grained sentiment recognition for multimodal information, 
the input image-text pairs are always complex. 
(1)
The semantics of sentence is complex which adds sentiment confusion among different aspects. 
Take Figure \ref{fig:intro} as an example, there are 3 aspects in the sentence with 3 different sentiments, 
%
The sentiment of ``mayor" can be easily confused by the keyword, ``Interesting'', which is of positive sentiment.
%
(2) 
The images contain a large amount of detailed information, and the visual contents are usually related to only one or several of the aspects.
 For example, 
as shown in Figure \ref{fig:intro}, 
the objects in red boxes are more helpful in analyzing the sentiment of ``Mayor Kadokawa'' than the other aspects.
The complex input greatly challenges the recognition of aspect-based sentiment.

Considering the multimodal input, existing methods are typically dedicated to associated visual and textual contents
\citep{juJointMultimodalAspectSentiment2021,lingVisionLanguagePreTrainingMultimodal2022a,yangCrossModalMultitaskTransformer2022}. 
\citet{juJointMultimodalAspectSentiment2021} uses image-text relation to evaluate the contribution of visual contents to aspect sentiment, based on which to determine whether the image is involved in sentiment analysis. 
\citet{lingVisionLanguagePreTrainingMultimodal2022a} and \citet{yangCrossModalMultitaskTransformer2022} align visual representations of objects and their attributes with corresponding textual contents.
%
To summarize, the whole image is directly associated with textual content in these methods.
%
%
Intuitively, without aligning image blocks to corresponding aspects, 
the coarse whole-image-text association can introduce aspect-irrelevant visual noise, which further hinders aspect sentiment analysis.
In addition, the performance can be further impacted by the textual noise from the confusion among different aspects.


In this paper, we propose an \underline{A}spect-\underline{o}riented \underline{M}ethod (\textbf{AoM}) to mitigate aforementioned noises from both image and text. 
%
AoM can detect aspect-relevant information from perspectives of both semantics and sentiment.
%
There are two key modules in AoM: Aspect-Aware Attention Module (A$^3$M) for semantically fine-grained image-text alignment and Aspect-Guided Graph Convolutional Network (AG-GCN) for sentiment information aggregation.
In A$^3$M, 
we first extract aspect features associated with each visual and textual token. 
And then aspect-relevant token representations are computed based on their relevance to the corresponding aspect features.
%
%
%
%
%
In AG-GCN, %
we first explicitly add sentiment embeddings to the
obtained representations of visual and textual tokens.
A multimodal weighted-association matrix is constructed containing aspect-to-image-block similarity and
word-to-word dependency. 
Then we use a graph convolutional network to aggregate sentiment information according to the constructed multimodal matrix.
%

The contributions can be summarized as follows:

 (1) We propose an aspect-oriented network to mitigate the visual and textual noises from the complex image-text interactions. 

(2) We design an aspect-aware attention module and an aspect-guided graph convolutional network to effectively detect aspect-relevant multimodal contents from the perspectives of semantic and sentiment, respectively.

(3) Experiments on two benchmark datasets, including Twitter2015 and Twitter2017, show that our approach generally outperforms the state-of-the-art methods.

\section{Related Work}

In this section, we review the existing methods for both ABSA and MABSA.

\subsection{Aspect-based Sentiment Analysis}


%
In the past few years, Aspect-Based Sentiment Analysis (ABSA) in the textual fields has attracted much attention and gained mature research \citep{chenRelationAwareCollaborativeLearning2020,ohDeepContextRelationAware2021, xuPositionAwareTaggingAspect2020a}. 
%
On the one hand, most recent works are based on the pre-trained language model BERT because of its remarkable performance in many NLP tasks \cite{liangAspectbasedSentimentAnalysis2022}.
On the other hand, 
some recent efforts focus on modeling the dependency relationship between aspects and their corresponding descriptions,  
in which graph convolutional networks (GCNs) \cite{chenEnhancedMultiChannelGraph2022,liangBiSynGATBiSyntaxAware2022a,liangJointlyLearningAspectFocused2020,liDualGraphConvolutional2021,pangDynamicMultiChannelGraph2021} or graph attention networks (GATs) \cite{yuanetal2020graph} over dependency with the syntactic structure of a sentence are fully exploited. 
%

\subsection{Multimodal Aspect-based Sentiment Analysis}
%
With the enrichment of multimodal users' posts in social media, researchers find that images offer great supplementary information in aspect term extraction \citep{10.1007/978-3-030-60450-9_12,zhang2018adaptive,Asgari_Chenaghlu_2021} and sentiment analysis \citep{wuSentimentWordAware2022,zhang2022learning,liCognitiveBrainModel2021,Hazarika2020,caiMultiModalSarcasmDetection2019}. 
Thus, Multimodal Aspect-based Sentiment Analysis (MABSA) begins to be widely studied.
MABSA task can be divided into two independent sub-tasks, i.e., Multimodal Aspect Term Extraction (MATE) and Multimodal Aspect-oriented Sentiment Classification (MASC). The former extracts all aspect terms in the sentence at the prompt of the image, and the latter predicts the sentiment polarities for the aspects.


\citet{juJointMultimodalAspectSentiment2021} first realizes MABSA in a unified framework and designs an auxiliary cross-modal relation detection to control whether the visual information will be used in prediction. 
%
For capturing cross-modal alignment, \citet{lingVisionLanguagePreTrainingMultimodal2022a} constructs a generative multimodal architecture based on BART for both vision-language pre-training and the downstream MABSA tasks. 
%
%
\citet{yangCrossModalMultitaskTransformer2022} dynamically controls the contributions of the visual information to different aspects via the trick that the lower confidence of the results predicted by purely textual is, the more contributions from images will be considered. 
%

On the one hand, the above methods ignore the alignment of fine-grained visual blocks and the corresponding aspects, which introduce irrelevant visual noise. 
%
On the other hand, modeling of syntax dependency and sentiment information for aspect descriptions is absent in these methods, which is proved important in sentiment analysis \citep{liangAspectbasedSentimentAnalysis2022,kalaivani2022senticnet,xu2022graph}. 


To tackle the aforementioned issues, we propose an aspect-oriented model consisting of Aspect-Aware Attention Module and Aspect-Guided Graph Convolutional Network which respectively work to capture semantic information by fine-grained image-text alignment and effectively aggregate aspect-relevant sentiment information.

\section{Methodology}
\subsection{Overview}

\noindent \textbf{Task Definition.} Formally, given a tweet that contains an image $V$ and a sentence with $n$ words $S=(w_1,w_2,...,w_n)$, our goal is to acquire the sequence $Y$ representing all aspects and their associated sentiment polarities. We formulate the output of MABSA as $Y=[a_1^s,a_1^e,s_1,...,a_i^s,a_i^e,s_i,...a_k^s,a_k^e,s_k]$, where $a_i^s$, $a_i^e$ and $s_i$ depict the start index, end index of the $i$-th aspect and its sentiment polarity in the tweet, and $k$ is the number of aspects.

\begin{figure*}[ht]
\centering 
\includegraphics[width = \linewidth]{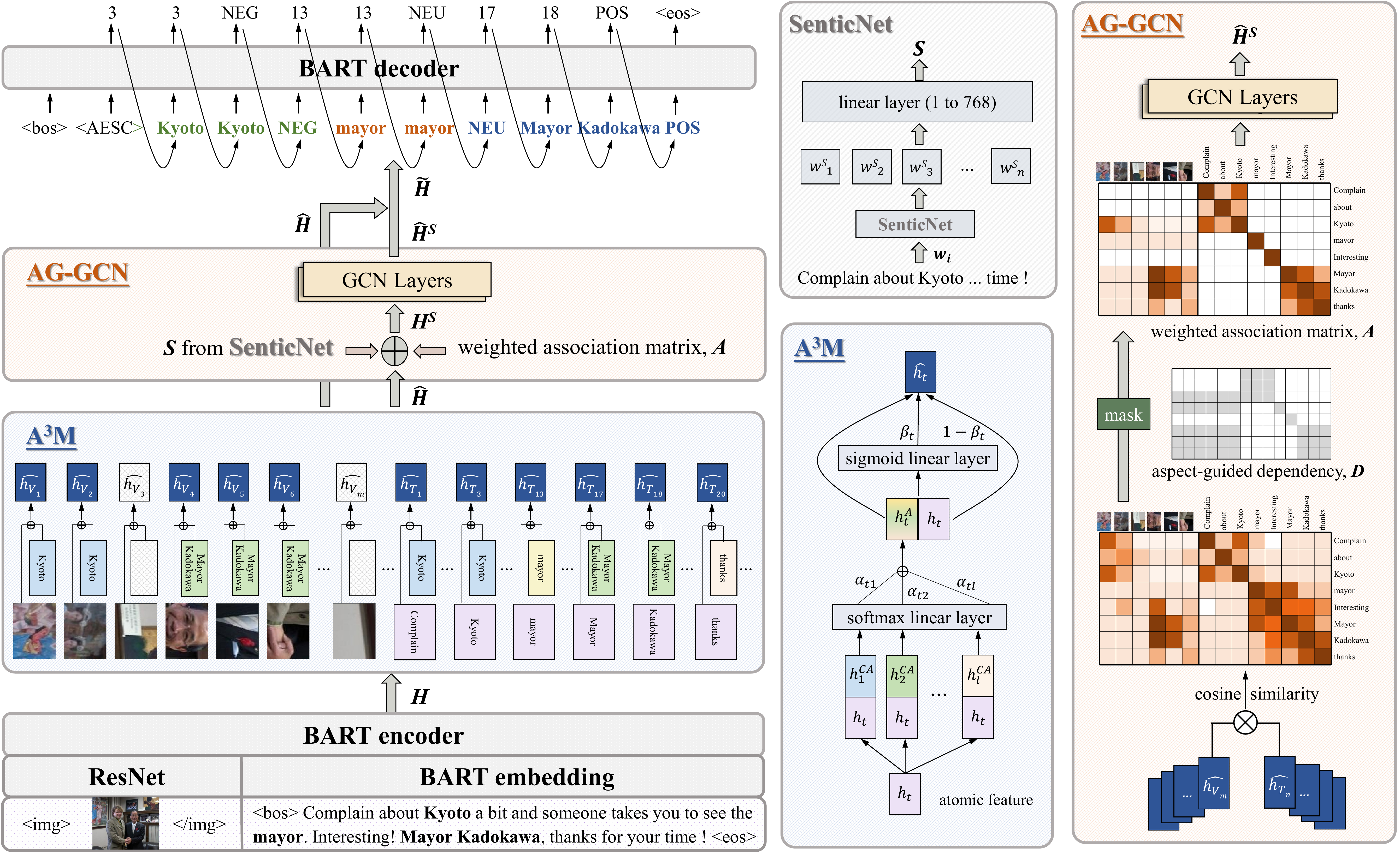}
\caption{The overview of our proposed aspect-oriented model AoM.}
\label{fig:arch}
\end{figure*}
%
\noindent \textbf{Model preview.} Figure \ref{fig:arch} shows the overview of our model architecture, which builds on an encoder-decoder architecture based on BART \citep{lewis2019bart}. Between the encoder and the decoder of BART, we creatively implement the Aspect-Aware Attention Module (A$^3$M) and Aspect-Guided Graph Convolutional Network (AG-GCN) to align the textual aspect to its associated visual blocks and textual description, simultaneously mitigate interference both from semantics and sentiment among different aspects. In the following subsections, we will illustrate the details of the proposed model.


\noindent \textbf{Feature Extractor.}
The initial word embeddings are obtained from the pre-trained BART due to its excellent ability of textual representation. The embeddings of visual blocks are obtained by pre-processing via ResNet \citep{ChenResnet2014} following \citep{yu2019entity}.
We consider every feature of a visual block or word token as an atomic feature. 
We add <img> and </img> before and after the visual features, <bos> and <eos> for the textual features.
Then, we concatenate the multimodal features as $X$ which is the input of BART encoder.

We can get the multimodal hidden state $H=\{h_0^V,...h_i^V,...h_m^V,h_0^T,...,h_j^T,...h_n^T\}$ with $m$ visual blocks and $n$ words, where $h_i^V$ and $h_j^T$ refer to features of the $i$-th visual block and the $j$-th word in the sentence. 
%

\subsection{Aspect-Aware Attention Module (A$^3$M)}

Since aspects are not specially modeled by BART encoder,
we creatively design the Aspect-Aware Attention Module (A$^3$M) aiming to capture aspect-relevant semantic information. For this purpose, we align the multimodal information of target objects and filter out the semantic noise from images. 

First, as aspects are usually noun phrases from the sentences, we extract those phrases as the \underline{c}andidate \underline{a}spects (CA) with the NLP tool Spacy\footnote{Spacy: https://spacy.io/}.
%
And from the hidden state $H$ of the BART encoder, we obtain the features of all candidate aspects denoted as $H^{CA}=\{h_1^{CA},...,h_i^{CA},...,h_l^{CA}\}$, where $l$ is the number of noun phrases in the sentence.
To get the relationship between candidate aspects and atomic features, we implement an attention-based mechanism guided by the candidate aspects.
Given the $t$-th hidden feature $h_t$, its attention distribution $\alpha_t$ over $k$ candidate aspects is obtained by: 
\begin{equation}
\resizebox{.8\hsize}{!}{$Z_t=tanh((W_{CA}H^{CA}+b_{CA})\oplus(W_Hh_t+b_H))$}, 
\end{equation}
\begin{equation}
\alpha_t=softmax(W_\alpha Z_t +b_\alpha), 
\end{equation}
where $Z_t \in \mathbb{R}^{2d\times k}$ is the comprehensive feature extracted from both the candidate aspects and the hidden states. $H^{CA} \in \mathbb{R}^{d\times k}$ denotes the features of candidate aspects. $W_{CA} \in \mathbb{R}^{d \times d}$, $W_H \in \mathbb{R}^{d\times d}$, $W_\alpha \in \mathbb{R}^{1\times 2d}$, $b_{CA}$, $b_H$ and $b_\alpha$ are the learned parameters.$\oplus$ is an operator between a matrix and a vector, where the vector is repeated into the appropriate size to concatenate with the matrix.
%
%
We then get the aspect-related hidden feature $h_t^A$ by calculating the weighted sum of all candidate aspects following the below equation:
\begin{equation}
h_t^A=\sum_i^k{\alpha_{t,i} h_i^{CA}}.
\end{equation}
%
For example, if a visual block is strongly associated with the $j$-th aspect,  the corresponding $\alpha_{t,j}$ is approximately 1. $h_t^A$ would be equal to the aspect semantically. And if the visual block is not related to any specific candidate aspects, 
both $\alpha_t$ and $h_t^A$ would be zero-like vectors of no information.

Considering that not every visual block can be used for prediction, $\beta_t$ is learned to add up the atomic feature $h_t$ and its aspect-related hidden feature $h^A_t$. Details are as follows:
\begin{equation}
\beta_t=sigmoid(W_{\beta}[W_1h_t;W_2h_t^A]+b_{\beta}), 
\end{equation}
\begin{equation}
\hat{h_t}=\beta_th_t+(1-\beta_t)h_t^A,
\end{equation}
where $W_{\beta}$, $W_1$, $W_2$, $b_{\beta}$ are parameters, and $[;]$ denotes the concatenation operator for vectors. $\hat{h_t} \in \hat{H}$ is the final output of A$^3$M after the semantic alignment and the noise reduction procedure.
Thus we get the noiseless and aligned information 
for every atomic feature. 



\begin{figure}
\centering 
\includegraphics[width = \linewidth]{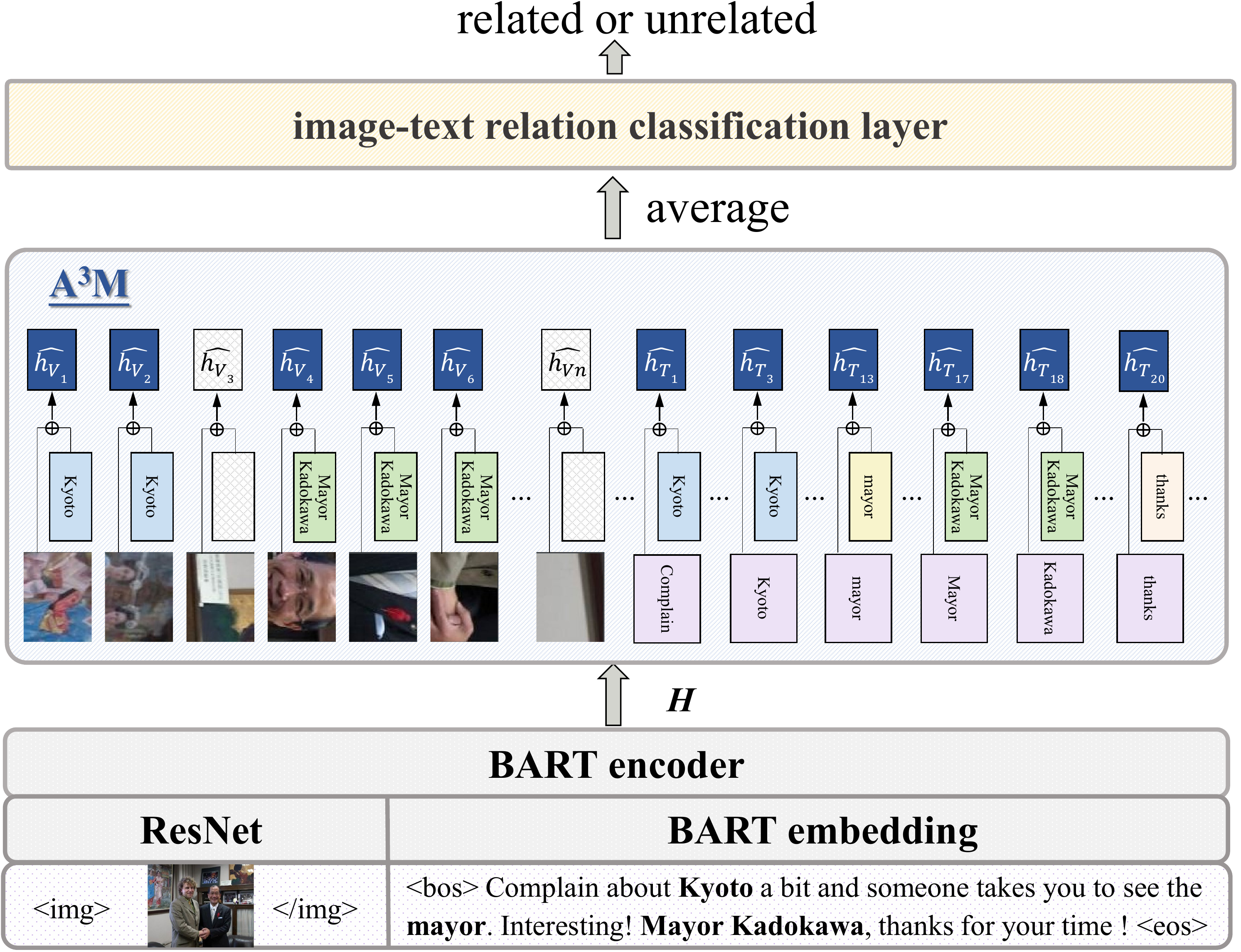}
\caption{The framework of the pre-training task.}
\label{fig:pretrain}
\end{figure}

\textbf{Pre-training} To align the two modalities and reduce noise, we conduct a pre-training task in A$^3$M. Specifically, we detect the image-text relationship on the datasets TRC \cite{vempala2019categorizing} as illustrated by Figure \ref{fig:pretrain}. 
We first obtain the average feature of image blocks from the output of A$^3$M and then pass it to a fully connected softmax layer, which outputs a probability distribution over whether the image is related to the text. 
%
%
Finally, we use cross entropy loss to train our model.

\subsection{Aspect-Guided Graph Convolutional Network (AG-GCN)}
The aspect-focused interaction between visual modality and textual modality in A$^3$M  concentrates on the context semantics, and that is not adequate for MABSA. Sentiment interference among different aspects still exists and influences sentiment prediction. Thus, we design the Aspect-Guided Graph Convolutional Network (AG-GCN) module to introduce external sentiment information and mitigate emotional confusion among different aspects to a certain extent. 


Specifically, for word $w_i$ in the sentence, we gain its affective score $w_i^S$ from SenticNet \citep{ma2018targeted} and project it to the space with the same dimension as 
$h_t^A$, with $s_i$ obtained. Then we add the sentiment feature $s_i$ to the output of A$^3$M: 
\begin{equation}
w_i^S=SenticNet(w_i),
\end{equation}
\begin{equation}
s_i=W_Sw_i^S+b_S,
\end{equation}
\begin{equation}
h_i^S=\hat{h_i}+s_i,
\label{HS}
\end{equation}

\noindent where $W_S$, $b_S$ are the learned parameters. $h_i^S$ is the feature with affective knowledge.

\begin{figure}
\centering 
\includegraphics[width = \linewidth]{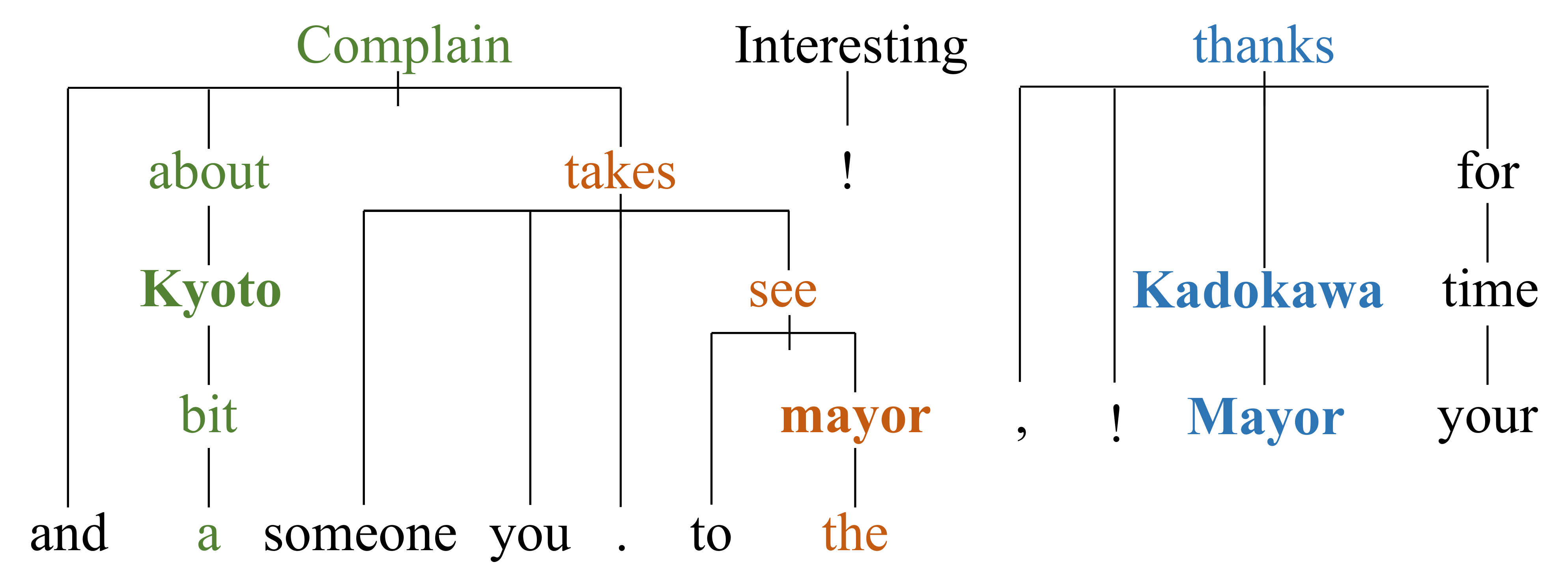}
\caption{The dependency tree of 
the example mentioned in the introduction.
}
\label{fig:dep}
\end{figure}

Next, we build a boolean dependency matrix $D$ among visual blocks and words. First, for the word-to-word part, submatrix $D_{TT}$ representing the dependency tree\footnote{We use spaCy toolkit to construct the dependency tree referring from https://spacy.io} of the input sentence like Figure \ref{fig:dep}. If two words can be associated within two generations, the element of $D_{TT}$ would be set to 1, otherwise 0 instead. For example, ``Kyoto'' is associated with  ``bit'' (child),``a'' (grandchild),``about'' (father) and ``Complain'' (grandfather). Second, the visual dependency submatrix $D_{VV}$ is initialized as a diagonal matrix. And as for the word-image-block dependency, denoted as $D_{TV}$ and equaled to $D_{VT}^T$, 
we set all the elements in the $i$-th line of $D_{TV}$ to 1 if the $i$-th word is an aspect, otherwise 0.
And the matrix $D$ 
is defined as:
\begin{equation}	
	\label{eq21}
	D = \left[
	\begin{matrix}
	D_{VV} &  D_{VT}\\ 
	D_{TV} &D_{TT}\\
	\end{matrix}
	\right],
\end{equation}  
%

 Considering the different importance of different dependencies, we attach weights onto $D$ with cosine similarity among $\hat{h_i}$ as follows:
 \begin{equation}
      A_{ij}=D_{ij}F_{cosine\_similarity}(\hat{h_i},\hat{h_j}),
 \end{equation}
 where both $D, A \in \mathbb{R}^{(m+n)\times(m+n)}$, and $A$ is the weighted association matrix. 
 %

AG-GCN takes $H^S$ from Eq.\ref{HS} as initial node representations in the graph. For the $i$-th node at the $l$-th layer, the hidden state $h_{i,l}^S$ is updated by the following equation:
\begin{equation}
   h_{i,l}^S=ReLU(\sum_{j=1}^nA_{ij}W_lh_{i,l-1}^S+b_{l}), 
\end{equation}
\noindent where $W_l$,$b_l$ are learned parameters and 
we use ReLU as activation function. Significantly, $h_{i,0}^S$ is equal to $h_i^S$. Accordingly, we get the final output $\hat{H}^S$ from the last GCN layer which is rich in sentiment information. Every underlying aspect aggregates its relevant information from both the image-text pair. Moreover, sentiment confusion of different aspects is weakened because the association matrix makes little interference among different aspects.

\subsection{Prediction and Loss Function}
The BART decoder takes the combination of $\hat{H}$, $\hat{H}^S$, and the previous decoder output $Y_{< t}$ as inputs, and predicts the token probability distribution as follows:

\begin{equation}
    \Tilde{H}=\lambda_1\hat{H}+\lambda_2\hat{H}^S,
\end{equation}
\begin{equation}
    h_t^d=Decoder(\Tilde{H};Y_{<t})
\end{equation}
\begin{equation}
    \overline{H}_T=(W+\Tilde{H}_T)/2
\end{equation}
\begin{equation}
P(y_t)=softmax([\overline{H}_T;C^d]h_t^d)
\end{equation}

\noindent where $\lambda_1$, $\lambda_2$ are the hyper-parameters to control the contribution from the two modules. $\Tilde{H}_T$ is the textual part of $\Tilde{H}$. $W$ denotes the embeddings of input tokens. $C^d$ means the embeddings of the [ positive, neutral, negative, <eos>]. The loss function is as follows:
\begin{equation}
    \mathcal{L}=-\mathbb{E}_{X\sim D}\sum_{t=1}^OlogP(y_t|Y_{<t},X),
\end{equation}
\noindent where $O=2M+2N+2$ is the length of Y, and X denotes the multimodal input.

\section{Experiment}
\subsection{Experimental settings}

%


\textbf{Datasets.}
Our two benchmark datasets are Twitter2015 and Twitter2017 (\citet{yu2019adapting}). As shown in the statistics of Table \ref{tab:twitter}, sentences with multiple aspects take up a considerable part of the two datasets. 
 




\begin{table}[t]
\centering
\resizebox{\linewidth}{!}{
\begin{tabular}{cccc}
\toprule
& Twitter2015& & Twitter2017\\

\midrule
\#sentence & 3,502 && 2,910\\
\#with one aspect & 2,159 (61.65\%) && 976 (33.54\%) \\
\#with multiple aspects &1,343 (38.35\%) && 1,934 (66.46\%)\\
\#with multiple sentiments & 1,257 && 1,690\\
\bottomrule
\end{tabular}
}
\caption{Statistics of the two benchmark datasets. Line 1 is the number of sentences. \#X in the last 3 lines denotes the number of sentences with such characteristics X.}
\label{tab:twitter}
\end{table}





\noindent \textbf{Implementation Details.}
Our model is based on BART \citep{lewis2019bart}, and the pre-training task is trained for 40 epochs with batch size 64, and for 35 epochs with batch size 16 on MABSA. The learning rates are both 7e-5 and hidden sizes are 768. Hyper-parameters $\lambda_1$ and $\lambda_2$ are 1 and 0.5 respectively. 
Besides, we pre-train A$^3$M on TRC dataset \cite{vempala2019categorizing},
which is divided into two groups according to whether the text is represented.

\noindent \textbf{Evaluation Metrics.} We evaluate the performance of our model on MABSA task and MATE task by Micro-F1 score (F1), Precision (P) and Recall (R), while on MASC task we use Accuracy (Acc) and F1 following previous studies.

\begin{table*}
\centering
\resizebox{0.9\linewidth}{!}{
\begin{tabular}{clccccccc}
\toprule
 & & \multicolumn{3}{c}{Twitter2015} & & \multicolumn{3}{c}{Twitter2017}  \\
\cline{3-5} \cline{7-9} 
& Methods & P & R & F1 & & P & R & F1 \\
\hline
\multirow{3}{*}{Text-based} & SPAN* \citep{hu2019open} &53.7 &53.9 &53.8 & & 59.6 &61.7 &60.6 \\
& D-GCN* \citep{chen2020joint} & 58.3& 58.8 &59.4 & &64.2 &64.1 & 64.1 \\
& BART* \citep{yan2021unified} & 62.9 & 65.0 & 63.9 & & 65.2 & 65.6 & 65.4 \\
\midrule
\multirow{9}{*}{Multimodal} & UMT+TomBERT* \citep{yu2020improving,yu2019adapting} & 58.4 &61.3 & 59.8 & & 62.3 & 62.4 &62.4 \\
& OSCGA+TomBERT* \citep{wu2020multimodal,yu2019adapting} & 61.7 & 63.4 & 62.5 & & 63.4 & 64.0 & 63.7 \\
& OSCGA-collapse* \citep{wu2020multimodal} & 63.1 &63.7&63.2 &&63.5&63.5&63.5\\
& RpBERT-collapse* \citep{sun2021rpbert} & 49.3 &46.9 &48.0& &57.0&55.4&56.2\\
& UMT-collapse \citep{yu2020improving} &61.0 &60.4 & 61.6 & & 60.8 & 60.0 &61.7 \\
&JML \citep{juJointMultimodalAspectSentiment2021}&65.0&63.2&64.1 & &66.5 & 65.5 & 66.0\\
& VLP-MABSA* \citep{lingVisionLanguagePreTrainingMultimodal2022a} &\underline{65.1} &68.3 & \underline{66.6} & & 66.9 & 69.2 & 68.0\\
& CMMT \citep{yangCrossModalMultitaskTransformer2022} & 64.6 &\underline{68.7} & 66.5 &  &\underline{67.6} & \underline{69.4} & \underline{68.5} \\
& AoM (ours) & \textbf{67.9}&  \textbf{69.3} & \textbf{68.6} &&  \textbf{68.4} & \textbf{71.0} & \textbf{69.7} \\
\bottomrule

\end{tabular}
}
\caption{Results of different methods for MABSA on the two Twitter datasets. * denotes the results from \citet{lingVisionLanguagePreTrainingMultimodal2022a}. The best results are bold-typed and the second best ones are underlined.}
\label{tab:MABSA_Results}
\end{table*}

\subsection{Baselines}
We compare our proposed model with four types of methods listed below.
%

\textbf{Approaches for textual ABSA.}
1) \textbf{SPAN} \citep{hu2019open} detects opinion targets with their sentiments. 2) \textbf{D-GCN} \citep{chen2020joint} models dependency relations among words via dependency tree. 3) \textbf{BART} \citep{yan2021unified}  solves seven ABSA subtasks in an end-to-end framework.

\textbf{Approaches for MATE.} 
1) \textbf{RAN} \citep{wu2020multimodal2} focus on alignment of text and object regions. 2) \textbf{UMT} \citep{yu2020improving} takes text-based entity span detection as an auxiliary task. 
3) \textbf{OSCGA} \citep{wu2020multimodal} 
foucus on alignments of visual objects and entities.

\textbf{Approaches for MASC.}
 1) \textbf{ESAFN} \citep{yu2019entity} is an entity-level sentiment analysis method based on LSTM. 2) \textbf{TomBERT} \citep{yu2019adapting} applies BERT to obtain aspect-sensitive textual representations. 3) \textbf{CapTrBERT} \citep{khan2021exploiting} translates images into text and construct an auxiliary sentence for fusion.

\textbf{Approaches for MABSA.} 1) \textbf{UMT-collapse} \citep{yu2020improving}, \textbf{OSCGA-collapse} \citep{wu2020multimodal} and \textbf{RpBERT-collapse} \citep{sun2021rpbert} are adapted from 
models for MATE
by using collapsed labels to represent aspect and sentiment pairs. 2) \textbf{UMT+TomBERT}, \textbf{OSCGA+TomBERT} are two pipeline approaches by combining UMT \citep{yu2020improving} or OSCGA \citep{wu2020multimodal} with TomBERT \citep{yu2019adapting}.  3) \textbf{JML} \citep{juJointMultimodalAspectSentiment2021} is the first joint model for MABSA with auxiliary cross-modal relation detection module.  
%
4) \textbf{CMMT} \citep{yangCrossModalMultitaskTransformer2022} implements a gate to control the multimodal information contributions during inter-modal interactions.
5) \textbf{VLP-MABSA} \citep{lingVisionLanguagePreTrainingMultimodal2022a} performs five task-specific pre-training tasks to model aspects, opinions and alignments.

\subsection{Main Results}
In this section, we show the excellent performance of AoM on the two datasets for the three tasks compared with SOTAs.

\textbf{Performance on MABSA:}
%
The results for MABSA are shown in Table \ref{tab:MABSA_Results}. \textbf{First}, our AoM far exceeds all text-based models, which means detection of richer visual information and textual information in our model is helpful. 
%
\textbf{Second}, multimodal pipeline methods and adaptive methods are generally unsatisfactory, because they ignore the interaction between the semantic information and sentiment for the two sub-tasks.
\textbf{Last}, AoM outperforms all multimodal methods in every metric. 
Especially, AoM achieves the improvement of 2\% and 1.2\% with respect to F1 in contrast with the second best models on two datasets (\textit{VLP-MABSA} for Twitter2015 and \textit{CMMT} for Twitter2017), 
which demonstrates the effectiveness of learning aspect-relevant visual blocks and textual words compared to focusing on all visual and textual inputs.
%


\begin{table}[t]
\centering
\resizebox{0.9\linewidth}{!}{
\begin{tabular}{lccccccc}
\toprule
 & \multicolumn{3}{c}{Twitter2015} & & \multicolumn{3}{c}{Twitter2017} \\
 \cline{2-4} \cline{6-8}
Methods&  P & R & F1 & & P & R & F1\\
\hline
RAN* &80.5 &81.5 &81.0 && 90.7 &90.7&90.0\\
UMT* &77.8 & 81.7 & 79.7&& 86.7 &86.8&86.7\\
OSCGA* & 81.7 & 82.1 & 81.9 && 90.2& 90.7 & 90.4\\
JML* &83.6 &81.2 & 82.4& &\underline{92.0} & 90.7 & 91.4 \\
VLP-MABSA* & 83.6 & \underline{87.9} &85.7& &90.8 & 92.6  & 91.7  \\
CMMT & \underline{83.9} &\textbf{88.1} &\underline{85.9} & & \textbf{92.2} &\textbf{93.9} &\textbf{93.1} \\
AoM (ours) & \textbf{84.6} & \underline{87.9} & \textbf{86.2}   & &  91.8  &  \underline{92.8} &  \underline{92.3} \\ 
\bottomrule
\end{tabular}
}
\caption{Results of different methods for MATE. * denotes the results from \citet{lingVisionLanguagePreTrainingMultimodal2022a}. }
\label{tab:MATE}
\end{table}

\textbf{Performance on MATE:}
%
 %
As shown in Table \ref{tab:MATE}, AoM is ahead of most of the current models and performs the best in Twitter 2015 by 0.3\% higher than the second best \textit{CMMT} on F1. The performance of \textit{CMMT} in Twitter2017 is 0.8\% higher than ours, probably due to our model wrongly predicting some noun phrases as aspects. 
%
But considering the improvement in MASC and MABSA, it is still worthy treating all noun phrases in the sentence as candidate aspects when acquiring aspect-relevant visual information.


\textbf{Performance on MASC:}
Table \ref{tab:MASC} shows the performance of MASC. It is exciting that our model outperforms the second-best results by 1.5\% and 2.6\% in accuracy, 2.1\% and 3.2\% points in F1 score on Twitter2015 and Twitter2017. 
%
%
It demonstrates that AoM has the ability to detect aspect-related sentiment information from both images and text, even disturbed by other noisy aspects.

\begin{table}
\centering
\resizebox{0.7\linewidth}{!}{
\begin{tabular}{lccccc}
\toprule
& \multicolumn{2}{c}{Twitter2015} & & \multicolumn{2}{c}{Twitter2017} \\
\cline{2-3} \cline{5-6}
Methods&  ACC & F1 & &  ACC & F1\\
\hline
ESAFN & 73.4 & 67.4 & &67.8 & 64.2\\
TomBERT& 77.2 &71.8 & & 70.5 & 68.0 \\
CapTrBERT & 78.0 & 73.2 &  &72.3 & 70.2 \\
JML & \underline{78.7} & - & & 72.7 & - \\
VLP-MABSA & 78.6 & \underline{73.8} & &\underline{73.8} &\underline{71.8} \\
CMMT & 77.9 & - & & \underline{73.8} & - \\
AoM (ours) & \textbf{80.2} & \textbf{75.9} & & \textbf{76.4} & \textbf{75.0} \\

\bottomrule
\end{tabular}
}
\caption{Results of different methods for MASC.}
\label{tab:MASC}
\end{table}

\subsection{Ablation Study}

In this section, we research the effectiveness of each component in AoM, the results are shown in Table \ref{tab:ablation}.

\begin{table}
\centering
\resizebox{\linewidth}{!}{
\begin{tabular}{lccccccc}
\toprule
 & \multicolumn{3}{c}{Twitter2015} & & \multicolumn{3}{c}{Twitter2017}  \\
\cline{2-4} \cline{6-8} 
Methods & P & R & F1 & & P & R & F1 \\
\hline
Full &  \textbf{67.9}&  69.3  & \textbf{68.6} &&  \textbf{68.4} & \textbf{71.0} & \textbf{69.7} \\
w/o A$^3$M\&AG-GCN & 65.7 & 67.3 & 66.5 && 66.5 & 69.0 & 67.8\\
w/o A$^3$M\&TRC & 62.1 & 61.0 & 61.6 & & 63.7 &64.1 & 63.9\\
w/o TRC & 66.8 & 68.4 & 67.6 & & 67.8 & 69.8 & 68.8\\
w/o AG-GCN & 67.0 & 69.4 & 68.2 & & 67.8 & 69.7 & 68.8 \\ 
w/o SenticNet & 65.7 & \textbf{70.5} & 68.0 && 68.1 & 69.4 & 68.7 \\
w/o TRC\&AG-GCN & 66.7 & 69.2 &68.0  & &  67.8 & 69.5 & 68.6\\

\bottomrule
\end{tabular}
}
\caption{The performance comparison of our full model and its ablated approaches.}
\label{tab:ablation}
\end{table}


\textbf{W/o A$^3$M\&AG-GCN} shows that after removing the two specially designed modules, the performance declines by 2.1\% on Twitter2015 and 1.9\% on Twitter2017. It fully demonstrates their contributions to learning effective information. 

\textbf{W/o A$^3$M\&TRC} performs worse after removing A$^3$M including the pre-training on TRC. It proves the necessity of modeling semantic alignment between visual blocks and aspects in A$^3$M. With the alignment, AG-GCN can obtain appropriate aspect-image-block and text-text association.

\begin{figure}
\centering 
\includegraphics[width = \linewidth]{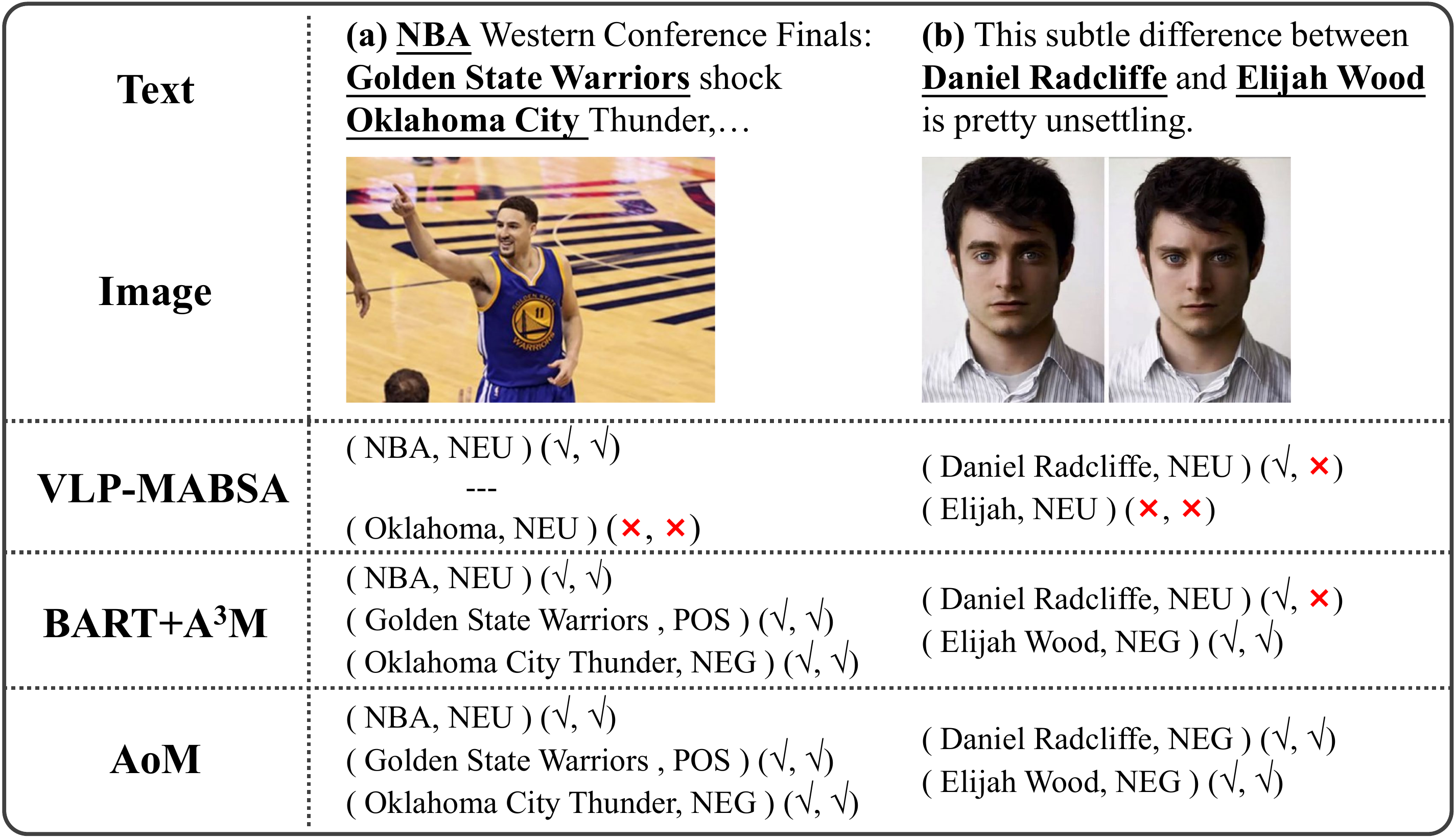}
\caption{Two cases with predictions by VLP-MABSA \citep{lingVisionLanguagePreTrainingMultimodal2022a}, BART+A$^3$M, and our model.}
\label{fig:case1}
\end{figure}

\textbf{W/o TRC pre-training} shows a slight drop after we remove the TRC pre-training on A$^3$M, which implies relevant pre-training task is useful for the model to learn better parameters.

\textbf{W/o AG-GCN} displays the performance without AG-GCN, declining by 0.42\% on Twitter2015 and 0.9\% on Twitter2017. It means that AG-GCN does make the prediction focus on specific aspects related to blocks and words with syntax dependencies. In other words, the multimodal interference from other aspects can be mitigated. 

\textbf{W/o SenticNet} is the model without sentiment information in AG-GCN. Its performance shows adding external affective knowledge can enhance the sentiment comprehension of the model.

\textbf{W/o TRC\&AG-GCN} is the BART model only with our A$^3$M module. 
We can see from Table \ref{tab:ablation} that\textit{ w/o TRC\&AG-GCN} improves \textit{w/o A$^3$M\&AG-GCN} by 1.5\% and 0.8\%. So it is effective to align the fine-grained visual block to related aspect and reduce irrelevant information.
%

\subsection{Case Study}
To better analyze how the Aspect-Aware Attention Module and Aspect-Guided Graph Convolutional Network work, we present the case study as follows.

Figure  \ref{fig:case1} displays two examples with predictions from VLP-MABSA \citep{lingVisionLanguagePreTrainingMultimodal2022a}, BART+A$^3$M and our AoM.
In example (a), VLP-MABSA misses the aspect ``Golden State Warriors'', gets an incomplete aspect ``Oklahoma City Thunder'' and wrongly predicts the sentiment. It may be caused by the interference from the visual region which represents pride expression of a person. However, BART+A$^3$M gets all right predictions due to the ability of aspect-oriented attention.
In example (b), compared with our whole model, BART+A$^3$M wrongly predicts the sentiment of ``Daniel Radcliffe'' which should be negative. 
We attribute the wrong prediction to lacking syntax association which benefits sentiment transmission. In other words, AG-GCN contributes to the correctness.

\subsection{Attention Visualization}

\begin{figure*}[ht]
\centering 
\includegraphics[width = \linewidth]{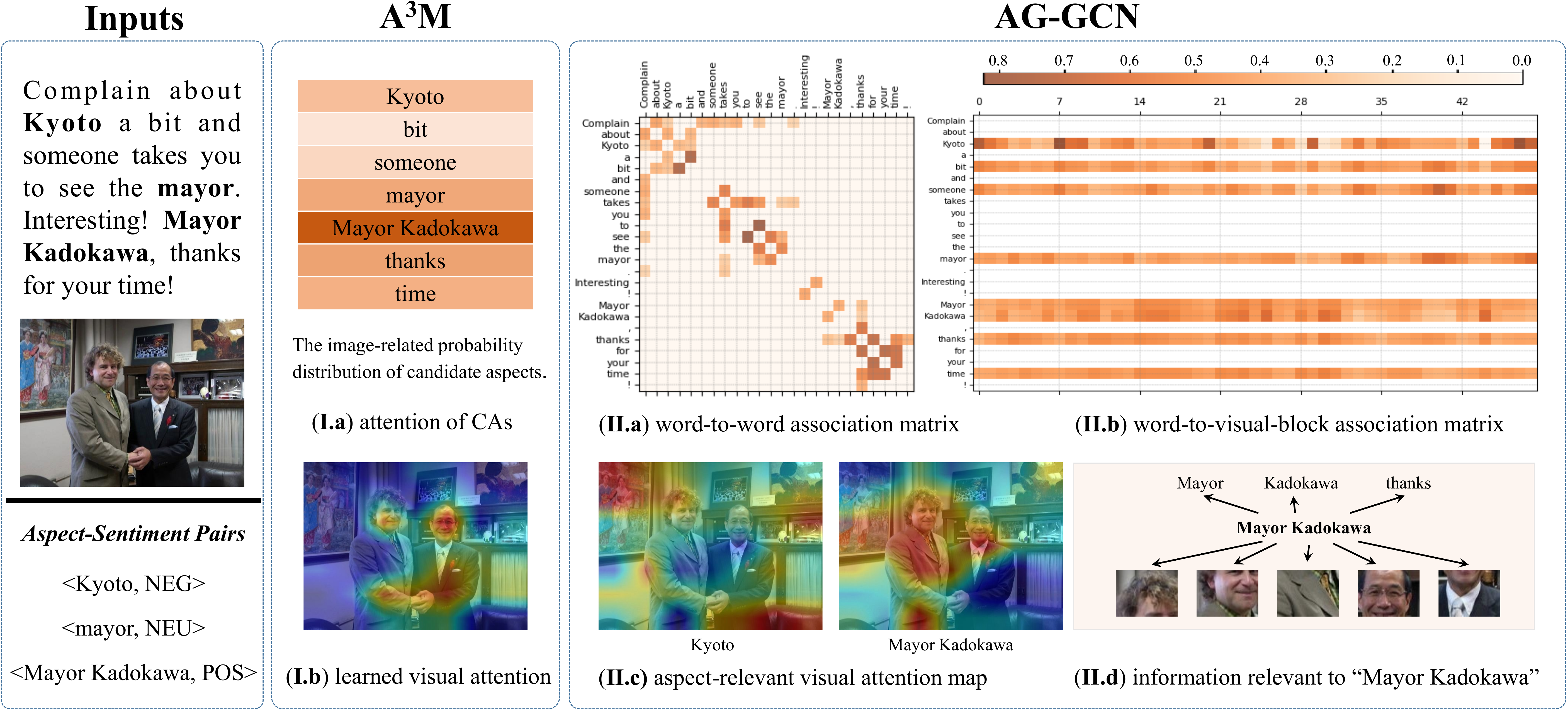}
\caption{Visualization of the attention maps in A$^3$M and the sub-parts of the weighted-association matrix AG-GCN.}
\label{fig:case}
\end{figure*}

To investigate the effectiveness of detecting aspect-relevant information, we visualize the attention process as shown in Figure \ref{fig:case}.

\textbf{For A$^3$M:}
(i) Figure \ref{fig:case}-(I.a) shows the attention weights of candidate aspects computed according to the images. 
We can see that ``Mayor Kadokawa'' is the most relevant aspect.
%
%
(ii) Figure \ref{fig:case}-(I.b) shows the proportions of the visual information reserved at the last step in A$^3$M, where we weighted add up the representations of visual blocks and the corresponding aspects. 
%
%
The heat map shows that the visual information associated with ``Mayor Kadokawa'' is reserved to a great extent, while the helpless information from other blocks is disregarded as noise. It demonstrates that attention in A$^3$M is able to detect aspect-relevant information.

\textbf{For AG-GCN:}
(i) Figure \ref{fig:case}-(II.a) shows the word-to-word part of the weighted association matrix. 
The matrix 
effectively excludes sentiment interference from other aspects by adding syntax dependency information. 
For example, the sentiment of ``mayor'' cannot be influenced by irrelevant keywords, such as ``Complain'' and ``thanks''. 
(ii) Figure \ref{fig:case}-(II.b) shows the dependencies between visual blocks and words. 
(iii) Specifically, we visualize the visual attention of aspects ``Kyoto'' (see Figure \ref{fig:case}-(II.c) left) and ``Mayor Kadokawa'' (see Figure \ref{fig:case}-(II.c) right).
We can see that ``Kyoto'' pays more attention to the pictures hanging on the wall which are full of Japanese elements related to the place, while ``Mayor Kadokawa'' focus more on the joyful expressions of the two people. 
(iv) Figure \ref{fig:case}-(II.d) shows the words and image blocks ``Mayor Kadokawa'' focused on in sentiment transmission. It's obvious that these attentions are helpful for the prediction.


\section{Conclusion}
In this paper, we proposed an aspect-oriented model (AoM) for the task of multimodal aspect-based sentiment analysis.
%
We use two specially designed modules to detect aspect-relevant information from the semantic and sentiment perspectives. 
On the one hand, to learn aspect-relevant semantic information especially from the image, we construct the Aspect-Aware Attention Module to align the visual information and descriptions to the corresponding aspect. 
%
On the other hand, to detect the aspect-relevant sentiment information, we explicitly add sentiment embedding into AoM. 
Then, a graph convolutional network is used to aggregate the semantic and sentiment embedding under the guidance of both image-text similarity and syntax dependency in sentences. 
The experimental results on two widely used datasets demonstrate the effectiveness of our method.


\section*{Limitations}
Though our proposed method outperforms current state-of-the-art methods, there are still many challenges we should overcome in future research. First,
for colloquial expression which confuses current dependency tree parser, we should come up with new solutions. Second, emotional prediction of tweet posts describing current issues needs external knowledge, which is absent in existing research.


\section*{Acknowledgments}
We thank anonymous reviewers for their valuable comments. This work was supported by the Natural Science Foundation of Tianjin, China (No.22JCJQJC00150, 22JCQNJC01580), the National Natural Science Foundation of China (No.62272250), Tianjin Research Innovation Project for Postgraduate Students (No.2022SKYZ232), and the Fundamental Research Funds for the Central Universities (No. 63231149).

\bibliography{custom}
\bibliographystyle{acl_natbib}

\end{document}